\documentclass[10pt,twocolumn,letterpaper]{article}

\usepackage[pagenumbers]{cvpr} 

\usepackage{CJKutf8}
\AtBeginDocument{\begin{CJK*}{UTF8}{bkai}}
\AtEndDocument{\end{CJK*}}

\definecolor{cvprblue}{rgb}{0.21,0.49,0.74}
\usepackage[pagebackref,breaklinks,colorlinks,allcolors=cvprblue]{hyperref}

\def\paperID{4} 
\def\confName{CVPR}
\def\confYear{2026}

\title{BST: Badminton Stroke-type Transformer for Skeleton-based Action Recognition in Racket Sports}

\author{Jing-Yuan Chang\\
National Tsing Hua University\\
\tt\small va6lue@gapp.nthu.edu.tw}

\begin{document}
\maketitle
\begin{abstract}
Badminton, known for having the fastest ball speeds among all sports, presents significant challenges to the field of computer vision, including player identification, court line detection, shuttlecock trajectory tracking, and player stroke-type classification.
In this paper, we introduce a novel video clipping strategy to extract frames of each player's racket swing in a badminton broadcast match.
These clipped frames are then processed by three existing models: one for Human Pose Estimation to obtain human skeletal joints, another for shuttlecock trajectory tracking, and the other for court line detection to determine player positions on the court.
Leveraging these data as inputs, we propose Badminton Stroke-type Transformer (BST) to classify player stroke-types in singles.
To the best of our knowledge, experimental results demonstrate that our method outperforms the previous state-of-the-art on the largest publicly available badminton video dataset (ShuttleSet), another badminton dataset (BadmintonDB), and a tennis dataset (TenniSet).
These results suggest that effectively leveraging ball trajectory is a promising direction for action recognition in racket sports.
The code is released at: \url{https://github.com/Va6lue/BST-Badminton-Stroke-type-Transformer}.
\end{abstract}
    
\section{Introduction}
\label{sec:intro}

In recent years, the rapid development of deep learning has catalyzed significant progress in sports analysis~\cite{BroadcastBadmintonAnalysis,MonoTrack,TennisCourtDetection,BasketballValidateOpenPose,BasketballActionPrediction,SoccerMap,BaseballDL}, aiming to provide athletes with objective statistical data to refine their techniques and devise effective strategies for continuous improvement.
Badminton, one of the world's most popular sports and known for having the fastest ball speeds among all sports, presents challenges for computer vision, including player identification, court line detection, shuttlecock trajectory tracking, and player stroke-type classification.

For the task of player stroke-type classification, models from the field of Human Action Recognition (HAR) can be leveraged.
The domain of HAR, which focuses on identifying the actions performed by individuals in videos, has evolved from directly analyzing RGB bounding boxes in the frames~\cite{HAR_RGB_1} to first extracting skeletal joints from these individuals and then analyzing them.
This extraction process is called Human Pose Estimation (HPE)~\cite{OpenPose_first,OpenPose_completeness,HRNet,RTMPose,MotionBERT}, and the remaining process is known as Skeleton-based Action Recognition (SAR)~\cite{ST-GCN,BlockGCN,SkateFormer,ProtoGCN,DeGCN,HDGCN,DG-STGCN,Hyperformer,FG-STFormer,2s-PGT}.
With this approach, we can effectively filter out extraneous background details and enable a concentrated focus on the nuances of human motion.
Nevertheless, even with these advanced SAR models~\cite{ST-GCN,BlockGCN,SkateFormer,ProtoGCN} (see \cref{subsec:SAR} for more details), achieving high accuracy in a broadcast badminton dataset faces several challenges.
These models are primarily designed to recognize human actions in everyday scenarios, where movements often exhibit significant variation, for instance, the contrast between the static action of drinking water and the dynamic motion of running.
However, players' racket-swinging actions can be categorized under a general "hitting" action in broader classifications, regardless of the specific stroke-type.
Furthermore, these actions are both rapid and brief, making it even more difficult to distinguish between different strokes.
Additionally, as noted in \cite{2s-PGT}, although some models support multi-person input, they do not fully take into account the relationship between interacting players.
These limitations underscore the challenge of distinguishing subtle differences in movements while utilizing limited information and determining which of the two players executed a particular stroke.

Fortunately, in badminton singles, the shuttlecock trajectory can serve as an interactive medium between the two players.
To the best of our knowledge, TemPose~\cite{TemPose} is the first model to incorporate shuttlecock trajectory information as an auxiliary input.
However, we argue that the shuttlecock trajectory plays an even more crucial role in recognizing badminton stroke actions.
Consider a hypothetical scenario where the shuttlecock is completely removed from a badminton match video, leaving only the two players performing air swings.
From a human perspective, would it still be possible to accurately determine the type of stroke performed based solely on the players' movements?
Moreover, players may deliberately perform deceptive movements to confuse their opponents, but the shuttlecock trajectory always reflects the actual stroke-type and contains no misleading information.
These points highlight the indispensable role of shuttlecock trajectory information in a robust badminton stroke recognition model.
Therefore, we treat the shuttlecock trajectory as a primary input in our model.
The main contributions of this paper are summarized as follows:

\begin{itemize}[leftmargin=2em]
    \item We introduce a novel badminton video clipping strategy to extract frames that have high relevance to the target badminton stroke swung by the player.
    \item We propose Badminton Stroke-type Transformer (BST) architecture, including its backbone BST-0 and several enhancing modules, that outperforms the previous state-of-the-art on the largest publicly available badminton video dataset (ShuttleSet~\cite{ShuttleSet}), another badminton dataset (BadmintonDB~\cite{BadmintonDB}), and a tennis dataset (TenniSet~\cite{TenniSet}).
    \item We show that \textit{effectively} leveraging ball trajectory information (when designing the method) can significantly improve the performance of badminton stroke classification, and even tennis stroke classification.
    \item Our BST series models achieve faster training speed compared to previous state-of-the-art methods, as detailed in the supplementary material.
    \item We present a detailed analysis and comparison of model performance using 2D and 3D joints in the supplementary material.
\end{itemize}

\section{Related Work}
\label{sec:related_work}

\subsection{Badminton Analysis System}
Anurag Ghosh \etal proposed an end-to-end framework~\cite{BroadcastBadmintonAnalysis} for automatic attribute tagging and analysis in badminton videos.
Building on this, Wei-Ting Chen \etal introduced a more advanced end-to-end analysis system~\cite{ExplorationBadminton}, which incorporates a visually appealing user interface for enhanced usability.
In the domain of stroke classification, Wei-Ta Chu and Samuel Situmeang developed a method~\cite{BadmintonVideoAnalysisBottomPlayer} specifically targeting the bottom player in singles matches.
Similarly, Yongkang Zhao proposed another stroke classification approach~\cite{AutoShuttlecockMotionRecognition} that also focuses on the bottom player, utilizing deep learning techniques to improve accuracy.
Magnus Ibh \etal introduced TemPose~\cite{TemPose} that can classify top and bottom player strokes in singles matches.
For serve detection, See Shin Yue \etal presented a specialized model~\cite{BadmintonServeCls} tailored for badminton.
Further advancing this area, the TrackNet series~\cite{TrackNet,TrackNetV2,TrackNetV3-Attention,TrackNetV3} primarily focuses on shuttlecock tracking and provides 2D trajectory predictions for downstream analysis systems.
Paul Liu and Jui-Hsien Wang proposed MonoTrack~\cite{MonoTrack}, an end-to-end system that not only tracks the shuttlecock but also extends its functionality by simulating 3D trajectories from 2D trajectory data.

\subsection{Skeleton-based Action Recognition (SAR)}
\label{subsec:SAR}
In this field, the development of Graph Convolutional Network (GCN)~\cite{GCN}-based models has been as rapid and competitive as that of Transformer~\cite{Transformer}-based models.

Sijie Yan \etal proposed ST-GCN~\cite{ST-GCN}, the first model to integrate GCN and Temporal Convolutional Network (TCN)~\cite{TCN} for SAR.
At the time, it was considered a groundbreaking development.

Yuxuan Zhou \etal introduced BlockGCN~\cite{BlockGCN}, which employs BlockGC (a custom-designed GCN) along with a multi-scale TCN.
Before applying these components, they incorporated Dynamic Topological Encoding.
Within BlockGC, they utilized a Static Topological Encoding based on the hop distance between joints and a learnable adjacency matrix to provide the model with greater flexibility in learning.
Remarkably, BlockGCN achieved state-of-the-art performance without relying on any attention mechanisms and was published at \textit{CVPR} 2024.

SkateFormer~\cite{SkateFormer}, proposed by Jeonghyeok Do and Munchurl Kim, builds upon the Transformer architecture with several modifications.
They integrated GCN, TCN, and their custom-designed Skate-MSA into the self-attention module.
The core idea of Skate-MSA is to transform the input data into four different types, which were constructed based on their observations of human motion patterns.
Each type is then processed separately. Their work was published at \textit{ECCV} 2024.

Hongda Liu \etal proposed ProtoGCN~\cite{ProtoGCN}, which is also based on a custom GCN and a multi-scale TCN.
Additionally, they incorporated Class-Specific Contrastive Loss to enhance model training.
Their approach achieved strong performance on several well-known public datasets in SAR and was later published at \textit{CVPR} 2025.

\section{Method}
\label{sec:method}

In this section, we present the key components of our method.
In \cref{subsec:vcs}, we present a video clipping strategy which segments a match video into clips.
In \cref{subsec:model_inputs}, we introduce a data preprocessing work which produces player poses, positions and shuttlecock trajectories from stroke clips.
In \cref{subsec:BST}, we present a Transformer-based architecture which takes those as inputs and produces classification results.

\subsection{Video Clipping Strategy}
\label{subsec:vcs}

\begin{figure*}[htbp]
    \centering
    \begin{minipage}[t]{0.48\linewidth}
        \centering
        \begin{subfigure}[t]{\linewidth}
            \centering
            \includegraphics[width=0.9\linewidth]{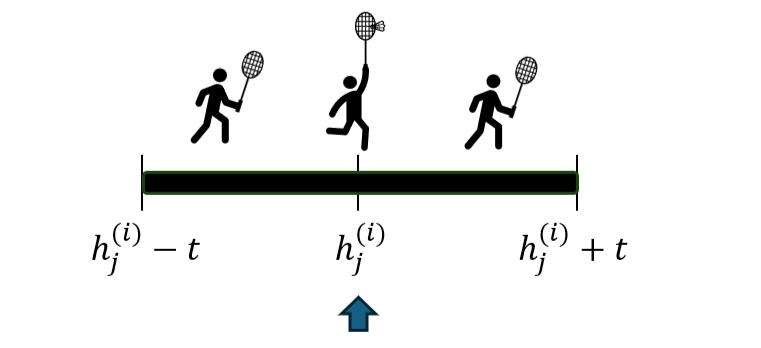}
            \caption{Fixed-width strategy}
            \label{fig:clipping_strategy-a}
        \end{subfigure}
        
        \vspace{0.3cm} 
        
        \begin{subfigure}[t]{\linewidth}
            \centering
            \includegraphics[width=0.9\linewidth]{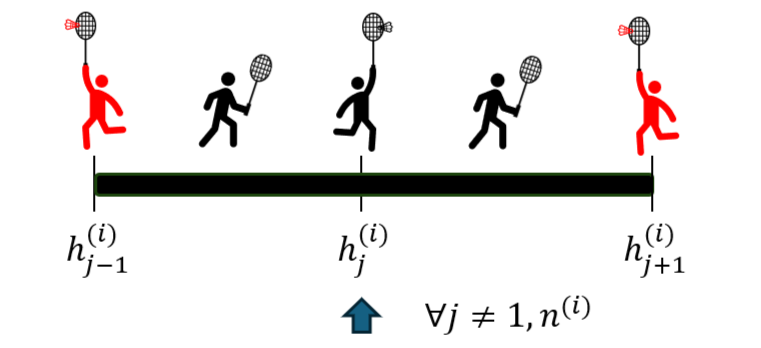}
            \caption{Complete-pose strategy}
            \label{fig:clipping_strategy-b}
        \end{subfigure}
    \end{minipage}
    \hfill
    \begin{minipage}[t]{0.48\linewidth}
        \centering
        \begin{subfigure}[t]{\linewidth}
            \centering
            \includegraphics[width=0.9\linewidth]{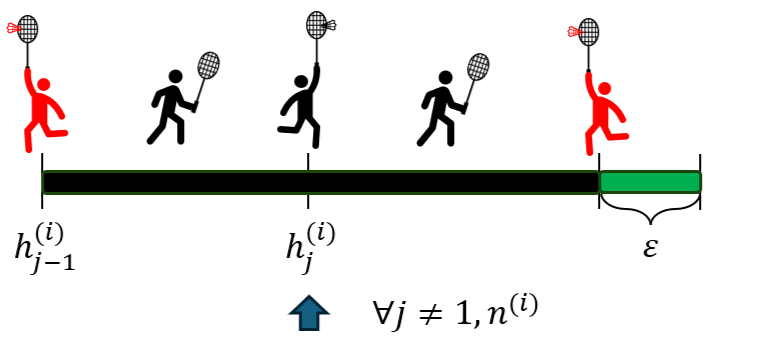}
            \caption{Our strategy (from the perspective of pose)}
            \label{fig:clipping_strategy-c}
        \end{subfigure}
        
        \vspace{0.3cm} 
        
        \begin{subfigure}[t]{\linewidth}
            \centering
            \includegraphics[width=0.9\linewidth]{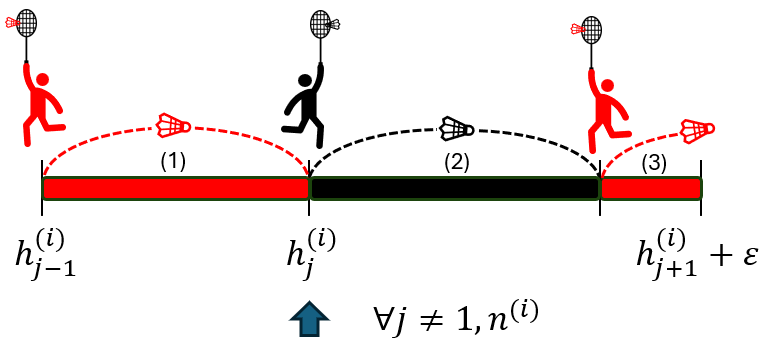}
            \caption{Our strategy (from the perspective of trajectory)}
            \label{fig:clipping_strategy-d}
        \end{subfigure}
    \end{minipage}
    
    \caption{
        The idea of video clipping strategies.
        The black player is the target player, and the red player is his/her opponent.
    }
    \label{fig:clipping_strategy}
\end{figure*}

In this subsection, we explain how we segment a match video into stroke clips.
There are many rallies in a match video.
A rally is a sequence of strokes that start from a serve and ends when the shuttlecock falls on the court and a point is awarded to a player.
A stroke refers to a badminton swing that hits the shuttlecock.
We first denote every rally $i$ that contains $n^{(i)}$ strokes $j$ to be a set like
\begin{equation}
    R_i=\{S^{(i)}_j\}^{n^{(i)}}_{j=1}=\{S^{(i)}_1, S^{(i)}_2,\ldots, S^{(i)}_{n^{(i)}}\},
\end{equation}
for $i=1, 2, \ldots, r$, where $r$ is the total number of rallies in the match video, and $j=1,2,\ldots,n^{(i)}$.
We now segment every rally into stroke clips.
Suppose a match video contains $N$ frames denoted by $F_1, F_2, \ldots, F_N$.
Some of the frames in the match video are hit frames, which are the frames in which the shuttlecock is in contact with, or closest to, the racket during a stroke.
Thus, every stroke $S^{(i)}_j$ in rally $R_i$ has a one-to-one correspondence with the hit frame number $h^{(i)}_j$.
A fixed-width clipping strategy similar to the approach used in \cite{AutoShuttlecockMotionRecognition} define the $j$-th stroke clip in rally $R_i$ to be
\begin{equation}
	C^{(i)}_j = \{F_{h^{(i)}_j - t}, F_{h^{(i)}_j - t + 1},\ldots, F_{h^{(i)}_j + t}\},
	\label{eq:fixed_width_clipping}
\end{equation}
for $j=1, 2, \ldots, n^{(i)}$ and $1\le i\le r$, as shown in \cref{sub@fig:clipping_strategy-a}.
The parameter, $t$, in this strategy ensures that the target hit frame $F_{h^{(i)}_j}$ is in the middle of the stroke clip $C^{(i)}_j$.
Ideally, $t$ is set to half the interval between two consecutive hit frames.
However, since $t$ is a fixed value, this presents a dilemma: a small value for $t$ may result in a clip that is too short, failing to capture complete player pose information, while a large value may include extraneous information (too many other hit frames) that is too far from the target stroke.
To address this issue, we need a different approach.
We can solve this by using the previous and next hit frame numbers from the opponent, as shown in \cref{sub@fig:clipping_strategy-b}.
Therefore, a complete-pose clip is obtained.

We further extend this approach.
Intuitively, consecutive strokes are highly correlated.
Therefore, our goal is for each stroke clip to contain information pertaining not only to the target stroke, but also to the opponent's previous and next strokes.
To achieve this, we propose our strategy as shown in \cref{sub@fig:clipping_strategy-c} and \cref{sub@fig:clipping_strategy-d}.
We just need to add a small value $\epsilon$, which is less than $t$, to the next hit frame number $h^{(i)}_{j+1}$ from the opponent.
As we mentioned in \cref{sec:intro}, the shuttlecock trajectory is an indispensable piece of information for stroke-type classification.
This small value, therefore, plays a crucial role in capturing partial trajectory for the opponent's next stroke, which is not included in the complete-pose strategy.
Comprehensively, these kinds of stroke clips can be thought of as three stages:

\begin{enumerate}[leftmargin=2em]
    \item Containing the second half of the pose and complete trajectory from the opponent's previous stroke, and the first half of the pose for the target stroke.
    \item Containing the second half of the pose and complete trajectory for the target stroke, and the first half of the pose from the opponent's next stroke.
    \item Containing the pose beginning in the second half of the opponent's next stroke, including its partial trajectory.
\end{enumerate}
Thus, our goal is achieved.
This allows the downstream model to not only judge based on the actual target trajectory in stage 2 and understand the target player's common strategies after going through stage 1, but also infer the target stroke-type back from stage 3.
The reason that stage 3 is shorter than stage 1, which means that $\epsilon$ cannot be set too large, is to preserve the characteristic of keeping the target hit frame close to the middle of the stroke clip under the ideal condition in the fixed-width strategy.
In other words, the complete target pose information should be placed in the central area of the stroke clip (\cref{sub@fig:clipping_strategy-c}).
Furthermore, the target trajectory information in stage 2 is also now nearly centrally located, as demonstrated in \cref{sub@fig:clipping_strategy-d}.
In formal terms, we can define
\begin{align}
	\hat{h}^{(i)}_{j-1} &= \begin{cases}
		h^{(i)}_{j-1}, & \text{if } j > 1
		\\
		h^{(i)}_j - t, & \text{if } j = 1,
	\end{cases}
	\nonumber
	\\
	\hat{h}^{(i)}_{j+1} &= \begin{cases}
		h^{(i)}_{j+1} + \epsilon, & \text{if } j < n^{(i)}
		\\
		h^{(i)}_j + t, & \text{if } j = n^{(i)},
		\label{eq:epsilon}
	\end{cases}
\end{align}
and a new stroke clip $\hat{C}^{(i)}_j$ in rally $R_i$ to be
\begin{equation}
	\hat{C}^{(i)}_j = \{F_{\hat{h}^{(i)}_{j-1}}, F_{\hat{h}^{(i)}_{j-1}+1},\ldots, F_{\hat{h}^{(i)}_{j+1}}\}.
	\label{eq:dynamic_clipping}
\end{equation}
If the target stroke being classified is the first or last stroke in a rally, we set the lower or upper bound of the stroke clip to be the same value as used in \cref{eq:fixed_width_clipping}.
More details are in the supplementary material.

\subsection{Model Inputs Extracted from Clips}
\label{subsec:model_inputs}

Now, we have a set of stroke clips extracted from the match videos.
The model inputs we need are player poses, shuttlecock trajectories, and player positions.

For the player poses and positions, we extract human poses from the clips using MMPose~\cite{MMPose} toolbox at first.
The models we utilized in MMPose are RTMPose~\cite{RTMPose} for 2D pose estimation and MotionBERT~\cite{MotionBERT} for further 3D pose estimation.
(We choose 2D poses for better downstream accuracy.)
However, there are more than two people, including the two players, spectators, and referees, in the clips.
To extract the exact two player poses, we need the court lines to determine the player positions to filter out the irrelevant poses.
The court lines information is available in one of our datasets.
For others not available, we used the algorithm proposed in MonoTrack~\cite{MonoTrack} or a deep learning model TennisCourtDetector~\cite{TennisCourtDetector} to extract the court lines.

For the shuttlecock trajectories, we utilize the 2D shuttlecock positions generated by TrackNetV3\footnote{There are two versions of TrackNetV3 developed by different authors. We choosed the one using attention for faster trajectory generating.} \cite{TrackNetV3-Attention,TrackNetV3} for each clip.
Despite the existence of the 3D shuttlecock trajectory physical model MonoTrack~\cite{MonoTrack}, the first work to estimate 3D shuttlecock trajectories over 2D, precise 3D estimation remains challenging.

\subsection{Badminton Stroke-type Transformer (BST)}
\label{subsec:BST}

\begin{figure*}[htbp]
    \centering
    \includegraphics[width=\linewidth]{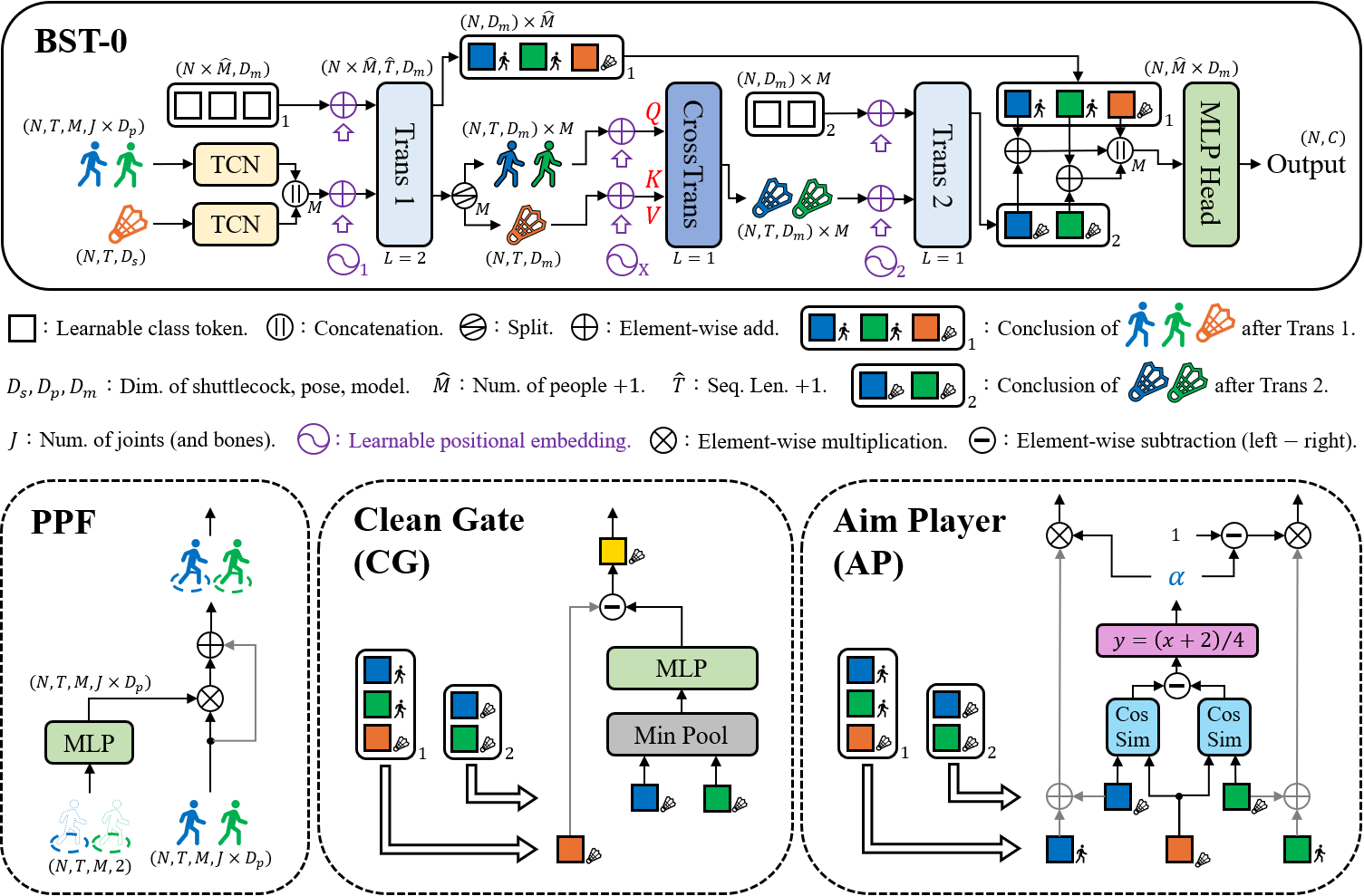}
    \caption{Architecture of BST. (In practice: Top player is blue, and Bottom player is green.)}
    \label{fig:BST}
\end{figure*}

The architecture of BST is derived from a Transformer-based model TemPose~\cite{TemPose}, which, to our knowledge, is the state-of-the-art badminton stroke classification model.
A key distinction is that BST shifts its focus from player poses to shuttlecock trajectory information.
The following subsections and \cref{fig:BST} describe the components of BST in detail.

\subsubsection{BST-0}
\label{subsubsec:BST-0}
BST-0 serves as the backbone of the BST architecture.
Initially, the player poses and shuttlecock positions are each processed through dedicated TCN modules.
After the outputs from this TCN are passed through a Transformer Encoder with two Transformer Layers, individual class tokens representing spatiotemporal features are obtained.

The Transformer Encoder modules in this paper are more similar to that used in TemPose~\cite{TemPose} than the original Transformer~\cite{Transformer}.
At the beginning, a learnable class token is prepended to the input sequence for conclusion, and a learnable positional embedding is added to the input as well.
We can formulate a Transformer Layer as follows:
\begin{align}
    \tilde{X}^{(l)} &= X^{(l-1)} + \text{MHSA}(\text{split}(\text{LN}(X^{(l-1)})))
    \\
    X^{(l)} &= \tilde{X}^{(l)} + \text{FFN}(\text{LN}(\tilde{X}^{(l)}))
\end{align}
where $X^{(l)} \in \mathbb{R}^{S \times D_{model}}$ is the output to the $l$-th layer in this Transformer Encoder, $S$ is the sequence length, $D_{model}$ is the dimension used in the model, $\text{LN}$ denotes the layer normalization, $\text{MHSA}$ is the multi-head self-attention mechanism, and $\text{FFN}$ is the feedforward network.
Before passing through the MHSA, the input $X^{(l-1)}$ which simplifies here to $X$ is split into:
\begin{align}
    Q &= \left[\ Q_1\ \cdots\ Q_{H}\ \right] = XW_Q \in \mathbb{R}^{S \times (D_A \times H)}
    \\
    K &= \left[\ K_1\ \cdots\ K_{H}\ \right] = XW_K \in \mathbb{R}^{S \times (D_A \times H)}
    \\
    V &= \left[\ V_1\ \cdots\ V_{H}\ \right] = XW_V \in \mathbb{R}^{S \times (D_A \times H)}
\end{align}
where $W_Q, W_K, W_V \in \mathbb{R}^{D_{model} \times (D_A \times H)}$ are the weight matrices, $D_A$ is the hyperparameter representing the dimension used in the attention modules, and $H$ is the number of the heads.
Further in the MHSA, we formulate as follows:
\begin{align}
    \text{MHSA}(Q, K, V) &= \left[\ \cdots\text{SA}(Q_h, K_h, V_h)\cdots\right]W_O
    \\
    \text{SA}(Q_h, K_h, V_h) &= softmax\left(\displaystyle\frac{Q_h(K_h)^\text{T}}{\sqrt{D_A}}\right)V_h
\end{align}
where $W_O \in \mathbb{R}^{(D_A \times H) \times D_{model}}$ represents the linear transformation from the attention dimension $D_A$ to the original model dimension $D_{model}$.

Besides the different module used on handling the poses at the beginning from TemPose~\cite{TemPose}, a key concept of our approach is ensuring that the model focuses on extracting critical information from the shuttlecock trajectory.
To achieve this, we design a Cross Transformer Layer (CrossTrans) with a Multi-Head Cross Attention (MHCA), which differs from MHSA in one key aspect: in the attention mechanism, $K$ and $V$ are derived from the shuttlecock trajectory latent, while $Q$ is generated from the player pose latent or the player position latent, respectively, to perform cross-attention.
As a result, the subsequent outputs are the shuttlecock trajectory latents belonging to the two players, respectively.
Then, both latents are fed into a second Transformer Encoder.
After that, the model leverages the overall shuttlecock information, the latent related to the blue player, and the latent related to the green player to generate logits through an MLP (multi-layer perceptron) Head.
Finally, a softmax function is applied to the outputs of the MLP Head to obtain probability values.

Trans1 and Trans2 are inspired by TemPose~\cite{TemPose} (Temporal and Interactional Transformer), but Trans2 is actually different from their Interactional Transformer that serves \textit{people} as a sequence.
Our Trans2 still processes the frame sequence to obtain the class token containing the trajectory information related to these two players, respectively.
For CrossTrans, our idea involves calculating the similarity between pose and trajectory, and using these similarity values to pick the value arrays in $V$ (from trajectory), which is a critical information from our perspective, an insight gained after observing our experimental results in \cref{sec:experiments} (TemPose-PF vs. TemPose-SF in \cref{tab:comparison_ShuttleSet_25_fixed_width_strategy}).

\subsubsection{Pose Position Fusion (PPF) Module}
This module generates a fused representation of pose and position.
An MLP within the module outputs coefficients, which are multiplied with the original poses.
This process imbues each joint with the influence of its corresponding court position.
The model, BST, is composed of BST-0 and this module.

\begin{table*}[htbp]
    \caption{
        Model comparison on ShuttleSet (25 classes) with fixed-width strategy.
        J and B in the modality denote the players' joints and bones.
        SP and PP represent additional inputs of shuttlecock and player positions.
        TemPose-PF and TemPose-SF are intermediate products derived by us from the original model, positioned between TemPose-V and TemPose-TF.
        The suffix "F" denotes the fusion of the additional inputs.
        \mbox{Acc-2} denotes top-2 accuracy.
    }
    \label{tab:comparison_ShuttleSet_25_fixed_width_strategy}
    \centering
    \begin{tabular}{l|c|ccc@{\hspace{0.5cm}}|@{\hspace{0.5cm}}cc@{\hspace{0.4cm}}c@{\hspace{0.6cm}}c}
        \toprule
        Model & Publication & Modality & SP & PP & Acc & Macro-F1 & Min-F1 & Acc-2 \\
        \midrule
        ST-GCN~\cite{ST-GCN} & \textit{AAAI} 2018 & J & \texttimes & \texttimes & 0.7758 & 0.7352 & 0.3726 & 0.9200 \\
        BlockGCN~\cite{BlockGCN} & \textit{CVPR} 2024 & J & \texttimes & \texttimes & 0.7652 & 0.7298 & 0.3788 & 0.9164 \\
        SkateFormer~\cite{SkateFormer} & \textit{ECCV} 2024 & J & \texttimes & \texttimes & 0.7710 & 0.7363 & 0.4355 & 0.8900 \\
        ProtoGCN~\cite{ProtoGCN} & \textit{CVPR} 2025 & J & \texttimes & \texttimes & 0.7746 & 0.7306 & 0.3326 & 0.9176 \\
        TemPose-V~\cite{TemPose} & \textit{CVPRW} 2023 & J & \texttimes & \texttimes & 0.7730 & 0.7360 & 0.4052 & 0.9248 \\
        \midrule
        TemPose-V~\cite{TemPose} & \textit{CVPRW} 2023 & J+B & \texttimes & \texttimes & 0.7756 & 0.7408 & 0.4286 & 0.9266 \\
        \midrule
        TemPose-PF &  & J+B & \texttimes & \checkmark & 0.7942 & 0.7704 & 0.4912 & 0.9384 \\
        \midrule
        TemPose-SF &  & J+B & \checkmark & \texttimes & 0.8100 & 0.7808 & 0.4872 & 0.9418 \\
        BST-0 &  & J+B & \checkmark & \texttimes & \textbf{0.8194} & \textbf{0.7924} & \textbf{0.5210} & \textbf{0.9444} \\
        \midrule
        TemPose-TF~\cite{TemPose} & \textit{CVPRW} 2023 & J+B & \checkmark & \checkmark & 0.8189 & 0.7943 & 0.4928 & 0.9496 \\
        BST &  & J+B & \checkmark & \checkmark & 0.8206 & 0.7952 & 0.5331 & 0.9499 \\
        BST-CG &  & J+B & \checkmark & \checkmark & 0.8210 & 0.7954 & 0.5296 & 0.9481 \\
        BST-AP &  & J+B & \checkmark & \checkmark & 0.8229 & \textbf{0.7992} & \textbf{0.5532} & 0.9484 \\
        BST-CG-AP &  & J+B & \checkmark & \checkmark & \textbf{0.8254} & 0.7983 & 0.5196 & \textbf{0.9503} \\
        \bottomrule
    \end{tabular}
\end{table*}

\begin{table*}[htbp]
    \caption{Model comparison on ShuttleSet (25 classes) with our clipping strategy.}
    \label{tab:comparison_ShuttleSet_25_our_strategy}
    \centering
    \begin{tabular}{l|ccc@{\hspace{0.5cm}}|@{\hspace{0.5cm}}cc@{\hspace{0.4cm}}c@{\hspace{0.6cm}}c}
        \toprule
        Model & Modality & SP & PP & Acc & Macro-F1 & Min-F1 & Acc-2 \\
        \midrule
        TemPose-SF & J+B & \checkmark & \texttimes & 0.8171 & 0.7936 & 0.5186 & 0.9515 \\
        BST-0 & J+B & \checkmark & \texttimes & \textbf{0.8284} & \textbf{0.8041} & \textbf{0.5822} & \textbf{0.9584} \\
        \midrule
        TemPose-TF~\cite{TemPose} & J+B & \checkmark & \checkmark & 0.8235 & 0.8028 & 0.5423 & 0.9567 \\
        BST & J+B & \checkmark & \checkmark & 0.8287 & 0.8049 & 0.5523 & \textbf{0.9602} \\
        BST-CG & J+B & \checkmark & \checkmark & 0.8307 & 0.8048 & 0.5444 & 0.9597 \\
        BST-AP & J+B & \checkmark & \checkmark & 0.8288 & 0.8056 & 0.5652 & 0.9586 \\
        BST-CG-AP & J+B & \checkmark & \checkmark & \textbf{0.8322} & \textbf{0.8097} & \textbf{0.5762} & 0.9594 \\
        \bottomrule
    \end{tabular}
\end{table*}

\subsubsection{Clean Gate (CG) Module}
This module is designed to refine the information derived from the shuttlecock trajectory by leveraging the interaction strength between the two players and the shuttlecock.
This helps the model filter out noise in the trajectory caused by the opponent's strokes.
The model, BST-CG, is composed of BST and this module, and the overall trajectory information before passing through MLP Head is replaced by the cleaned trajectory information after passing through this module.

\subsubsection{Aim Player (AP) Module}
We compare the cosine similarity between the overall trajectory information and each player's trajectory information.
The difference between these two values yields $\tilde{\alpha}$.
Since cosine similarity ranges from -1 to 1, $\tilde{\alpha} \in [-2, 2]$.
We normalize it using $\alpha = (\tilde{\alpha} + 2) / 4$ so that $\alpha \in [0, 1]$.
$\alpha$ denotes the coefficient on the blue player's information, so $1-\alpha$ is the coefficient on the green player's information.
This ensures that the player with a stronger correlation to the overall shuttlecock trajectory exerts a greater influence on the input to the MLP Head.
The model, BST-AP, is composed of BST and this module but without passing the overall trajectory information through MLP Head.
This is because we have a final model, BST-CG-AP, which combines both CG and AP modules with BST.

\section{Experiments}
\label{sec:experiments}
In this section, we present the three datasets used in our experiments and compare our results with those obtained by several existing methods, including ST-GCN~\cite{ST-GCN}, BlockGCN~\cite{BlockGCN}, SkateFormer~\cite{SkateFormer}, ProtoGCN~\cite{ProtoGCN}, and TemPose~\cite{TemPose}.

\begin{table*}[htbp]
    \caption{Model comparison trained on 25\% ShuttleSet (25 classes) with fixed-width strategy.}
    \label{tab:comparison_ShuttleSet_25_reduced_training_set}
    \centering
    \begin{tabular}{l|ccc@{\hspace{0.5cm}}|@{\hspace{0.5cm}}ccc}
        \toprule
        Model & Modality & SP & PP & Acc & Macro-F1 & Acc-2 \\
        \midrule
        TemPose-SF & J+B & \checkmark & \texttimes & 0.6504 & 0.5518 & 0.8262 \\
        BST-0 & J+B & \checkmark & \texttimes & \textbf{0.6946} & \textbf{0.6248} & \textbf{0.8566} \\
        \midrule
        TemPose-TF~\cite{TemPose} & J+B & \checkmark & \checkmark & 0.6636 & 0.5904 & 0.8344 \\
        BST & J+B & \checkmark & \checkmark & 0.6884 & 0.6196 & 0.8552 \\
        BST-CG & J+B & \checkmark & \checkmark & 0.7020 & \textbf{0.6334} & \textbf{0.8672} \\
        BST-AP & J+B & \checkmark & \checkmark & \textbf{0.7038} & 0.6302 & 0.8584 \\
        BST-CG-AP & J+B & \checkmark & \checkmark & 0.6924 & 0.6238 & 0.8566 \\
        \bottomrule
    \end{tabular}
\end{table*}

\begin{table*}[htbp]
    \caption{
        Model comparison on BadmintonDB (18 classes) with predefined strategy.
    }
    \label{tab:comparison_BadmintonDB}
    \centering
    \begin{tabular}{l|c|ccc@{\hspace{0.5cm}}|@{\hspace{0.5cm}}ccc}
        \toprule
        Model & Publication & Modality & SP & PP & Acc & Macro-F1 & Acc-2 \\
        \midrule
        ST-GCN~\cite{ST-GCN} & \textit{AAAI} 2018 & J & \texttimes & \texttimes & 0.5602 & 0.4870 & 0.7732 \\
        BlockGCN~\cite{BlockGCN} & \textit{CVPR} 2024 & J & \texttimes & \texttimes & 0.5782 & 0.4958 & 0.7970 \\
        ProtoGCN~\cite{ProtoGCN} & \textit{CVPR} 2025 & J & \texttimes & \texttimes & 0.5656 & 0.4798 & 0.7888 \\
        TemPose-V~\cite{TemPose} & \textit{CVPRW} 2023 & J & \texttimes & \texttimes & 0.5708 & 0.5248 & 0.8134 \\
        \midrule
        TemPose-V~\cite{TemPose} & \textit{CVPRW} 2023 & J+B & \texttimes & \texttimes & 0.5764 & 0.5138 & 0.8308 \\
        \midrule
        TemPose-PF &  & J+B & \texttimes & \checkmark & 0.5986 & 0.5516 & 0.8536 \\
        \midrule
        TemPose-SF &  & J+B & \checkmark & \texttimes & 0.6204 & 0.5440 & 0.8380 \\
        BST-0 &  & J+B & \checkmark & \texttimes & \textbf{0.6460} & \textbf{0.5671} & \textbf{0.8569} \\
        \midrule
        TemPose-TF~\cite{TemPose} & \textit{CVPRW} 2023 & J+B & \checkmark & \checkmark & 0.6199 & 0.5448 & 0.8517 \\
        BST &  & J+B & \checkmark & \checkmark & 0.6420 & 0.5600 & 0.8620 \\
        BST-CG &  & J+B & \checkmark & \checkmark & 0.6437 & 0.5734 & \textbf{0.8639} \\
        BST-AP &  & J+B & \checkmark & \checkmark & \textbf{0.6517} & \textbf{0.5799} & 0.8628 \\
        BST-CG-AP &  & J+B & \checkmark & \checkmark & 0.6392 & 0.5580 & 0.8574 \\
        \bottomrule
    \end{tabular}
\end{table*}

\begin{table*}[htbp]
    \caption{
        Model comparison on TenniSet (6 classes) with predefined strategy.
    }
    \label{tab:comparison_TenniSet}
    \centering
    \begin{tabular}{l|c|ccc@{\hspace{0.5cm}}|@{\hspace{0.5cm}}cc@{\hspace{0.4cm}}c@{\hspace{0.6cm}}c}
        \toprule
        Model & Publication & Modality & SP & PP & Acc & Macro-F1 & Min-F1 & Acc-2 \\
        \midrule
        ST-GCN~\cite{ST-GCN} & \textit{AAAI} 2018 & J & \texttimes & \texttimes & 0.9890 & 0.9882 & 0.9766 & 0.9956 \\
        TemPose-V~\cite{TemPose} & \textit{CVPRW} 2023 & J & \texttimes & \texttimes & 0.9824 & 0.9822 & 0.9732 & 0.9946 \\
        \midrule
        TemPose-V~\cite{TemPose} & \textit{CVPRW} 2023 & J+B & \texttimes & \texttimes & 0.9780 & 0.9770 & 0.9622 & 0.9922 \\
        \midrule
        TemPose-PF &  & J+B & \texttimes & \checkmark & 0.9844 & 0.9846 & 0.9688 & 0.9948 \\
        \midrule
        TemPose-SF &  & J+B & \checkmark & \texttimes & 0.9766 & 0.9765 & 0.9608 & 0.9938 \\
        BST-0 &  & J+B & \checkmark & \texttimes & \textbf{0.9904} & \textbf{0.9901} & \textbf{0.9807} & \textbf{0.9992} \\
        \midrule
        TemPose-TF~\cite{TemPose} & \textit{CVPRW} 2023 & J+B & \checkmark & \checkmark & 0.9866 & 0.9872 & 0.9767 & 0.9956 \\
        BST &  & J+B & \checkmark & \checkmark & 0.9890 & 0.9889 & 0.9795 & 0.9978 \\
        BST-CG &  & J+B & \checkmark & \checkmark & 0.9913 & 0.9913 & 0.9829 & 0.9986 \\
        BST-AP &  & J+B & \checkmark & \checkmark & \textbf{0.9923} & \textbf{0.9922} & \textbf{0.9845} & 0.9990 \\
        BST-CG-AP &  & J+B & \checkmark & \checkmark & 0.9878 & 0.9879 & 0.9783 & 0.9970 \\
        \bottomrule
    \end{tabular}
\end{table*}

\subsection{Datasets}
\label{subsec:datasets}

\textbf{ShuttleSet}\footnote{
    ShuttleSet has labels and video links.
    These videos are on BWF TV (\url{https://www.youtube.com/@bwftv}).
}~\cite{ShuttleSet}, the largest publicly available badminton video dataset, contains 104 sets, 3,685 rallies, and 36,492 strokes across 44 matches played between 2018 and 2021, featuring 27 top-ranking male and female singles players.
After excluding incorrectly labeled and problematic data, a total of 40 matches, 33481 strokes remained for analysis.
We set 30 matches for training, 5 matches for validation, and 5 matches for testing.

\vspace{0.8em}

\noindent\textbf{BadmintonDB}~\cite{BadmintonDB} contains 7658 strokes across 8 badminton singles matches after data cleaning.
These matches were played by two players, Kento Momota and Anthony Sinisuka Ginting, in 2018 and 2019.
The time period of each clip is predefined, and on our observation, the target hit frames are always near the end of the clips (about half of them even after the end).
This means that, usually, the information in stage 1 of our clipping strategy is utilized only.
We split the dataset into 8:1:1 for training, validation, and testing on each class.

\vspace{0.8em}

\noindent\textbf{TenniSet}~\cite{TenniSet} contains 3569 strokes across 5 tennis singles matches.
The time period of each clip is also predefined.
We followed the split configuration predefined in this dataset for training, validation, and testing.

\subsection{Results}

\textbf{Results on ShuttleSet (merged).}\footnote{
    Several stroke-types are merged. Refer to the supplementary material for details.
}\quad\cref{tab:comparison_ShuttleSet_25_fixed_width_strategy} demonstrates that even the most advanced SAR models struggle to achieve satisfactory accuracy. Without incorporating additional inputs, such as shuttlecock trajectories and player positions, significant performance improvements are difficult to attain.
However, integrating such information is not straightforward for these models.
This highlights the need for specialized models like TemPose~\cite{TemPose}, which are specifically designed to handle these additional inputs.

Initially, by focusing solely on player position fusion, TemPose-PF achieves results that surpass those of TemPose-V.
The reason for this improvement is that player positions imply some information about the shuttlecock trajectory, that is, the players are always chasing the shuttlecock during a rally.
On the other hand, integrating only shuttlecock positions, TemPose-SF achieves a greater improvement over TemPose-V than TemPose-PF does.
Furthermore, combining both types of fusion further enhances performance.
Notably, our BST series achieves some improvement over TemPose on 18/20 metrics when using the fixed-width clipping strategy.

Moreover, our clipping strategy yields even greater performance gains for all TemPose and BST series models, as shown in \cref{tab:comparison_ShuttleSet_25_our_strategy}.

Finally, we also conducted experiments using only (each class) 25\% of the training set and tested on the same testing set, as shown in \cref{tab:comparison_ShuttleSet_25_reduced_training_set}.
The results indicate that our BST architecture demonstrates superior generalization capability compared to TemPose when the training set is small and reference information is limited.

\vspace{0.8em}

\noindent\textbf{Results on BadmintonDB.}\quad As demonstrated in \cref{tab:comparison_BadmintonDB}, these advanced SAR models still struggle to achieve satisfactory accuracy.
While the TemPose-SF outperforms TemPose-PF on accuracy, it falls short on macro-F1 and top-2 accuracy.
Furthermore, we found that building upon an SP with the addition of PP resulted in performance drops on several metrics (\eg TemPose-SF vs. TemPose-TF and BST-0 vs. BST).
Nevertheless, our BST series consistently outperforms previous methods.

\vspace{0.8em}

\noindent\textbf{Results on TenniSet.}\quad This dataset is less challenging than the previous two, primarily due to its reduced number of categories with a more balanced distribution.
As shown in \cref{tab:comparison_TenniSet}, all models achieve very high accuracy, macro-F1, top-2 accuracy, and even min-F1.
While adding bones information slightly yields a modest improvement on the previous two datasets, it results in a slight performance drop on this dataset.
Despite this, we still used the J+B modality on the subsequent models for consistency.
This time, ST-GCN outperforms TemPose models on most metrics.
However, our BST series still achieves the best results across most metrics.

\section{Conclusion}
\label{sec:conclusion}

In this paper, we first introduce a novel clipping strategy that segments a match video into clips containing highly relevant frames for the target stroke.
We propose BST architecture, including its backbone BST-0 and several enhancing modules, for badminton stroke classification.
Through experiments, we show that our method outperforms existing state-of-the-art methods not only in two badminton datasets but also in a tennis dataset, highlighting the importance of the ball information.
A detailed comparison of model training speed is provided in the supplementary material, along with a comprehensive analysis of performance using 2D vs. 3D joints.

{
    \small
    \bibliographystyle{ieeenat_fullname}
    \bibliography{main}
}

\clearpage
\appendix
\setcounter{page}{1}
\maketitlesupplementary

\renewcommand{\thetable}{\Alph{table}}
\renewcommand{\thefigure}{\Alph{figure}}
\setcounter{table}{0}
\setcounter{figure}{0}

\section*{Contents}
\begin{enumerate}[leftmargin=2em, label={\Alph*.}, itemsep=0.6em]
    \item \hyperlink{training_speed}{Training Speed}
    \item \hyperlink{discussion}{Discussion}\vspace{0.4em}
    \begin{enumerate}[label={\arabic*.}, itemsep=0.4em]
        \item \hyperlink{discussion_cls}{Classifying Difficulty on ShuttleSet}
        \item \hyperlink{discussion_2D_vs_3D}{Player 2D Joints vs. 3D Joints}
        \item \hyperlink{discussion_future_work}{Limitations and Future Work}
    \end{enumerate}
    \item \hyperlink{clipping_details}{Details in Clipping}\vspace{0.4em}
    \begin{enumerate}[label={\arabic*.}, itemsep=0.4em]
        \item Source of the Hit Frame and Rally Information
        \item Practical Settings
    \end{enumerate}
    \item \hyperlink{training_settings}{Training Settings}
    \item \hyperlink{datasets_details}{Details of the Datasets}
\end{enumerate}

\hypertarget{training_speed}{}
\section{Training Speed}
On a NVIDIA RTX 4090 setup, BST-CG-AP trains slightly faster than TemPose-TF, as shown in \cref{tab:training_speed}.

\begin{table}[htbp]
    \caption{
        Model training speed per epoch on ShuttleSet and the numbers of their parameters.
        TemPose-TF* is the original implementation version.
    }
    \label{tab:training_speed}
    \centering
    \begin{tabular}{l|crc}
        \toprule
        Model & Seq Len & Ep. Speed & \#Param \\
        \midrule
        ST-GCN~\cite{ST-GCN} & 30 & 11.1 s & 3.08M \\
        BlockGCN~\cite{BlockGCN} & 30 & 111.6 s & 1.50M \\
        SkateFormer~\cite{SkateFormer} & 30 & 41.8 s & 2.38M \\
        ProtoGCN~\cite{ProtoGCN} & 30 & 42.4 s & 4.11M \\
        \midrule
        TemPose-TF*~\cite{TemPose} & 100 & 11.0 s & 1.71M \\
        TemPose-TF~\cite{TemPose} & 100 & 10.0 s & 1.74M \\
        BST-CG-AP & 100 & 9.7 s & 1.87M \\
        \bottomrule
    \end{tabular}
\end{table}

\begin{figure}[htbp]
    \centering
    \includegraphics[width=\linewidth]{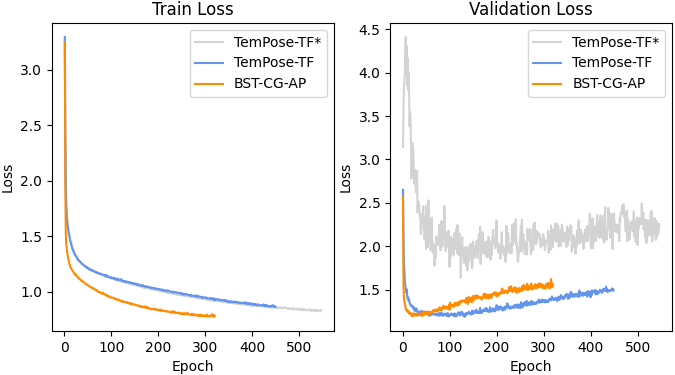}
    \caption{
        Loss Curves of BST-CG-AP, TemPose-TF and TemPose-TF*.
    }
    \label{fig:loss_curves}
\end{figure}

\cref{fig:loss_curves} also shows that training BST-CG-AP converges faster than TemPose-TF.
The performance gap between TemPose-TF* and TemPose-TF is likely due to the authors' use of a single TCN object to handle both player positions and the shuttlecock.

\hypertarget{discussion}{}
\section{Discussion}
\label{sec:discussion}

\begin{table*}[htbp]
    \caption{
        Model comparison on ShuttleSet (35 classes) with fixed-width strategy.
    }
    \label{tab:comparison_ShuttleSet_35_fixed_width_strategy}
    \centering
    \begin{tabular}{l|c|ccc@{\hspace{0.5cm}}|@{\hspace{0.5cm}}ccc}
        \toprule
        Model & Publication & Modality & SP & PP & Acc & Macro-F1 & Acc-2 \\
        \midrule
        ST-GCN~\cite{ST-GCN} & \textit{AAAI} 2018 & J & \texttimes & \texttimes & 0.7198 & 0.6396 & 0.8890 \\
        BlockGCN~\cite{BlockGCN} & \textit{CVPR} 2024 & J & \texttimes & \texttimes & 0.7108 & 0.6310 & 0.8822 \\
        SkateFormer~\cite{SkateFormer} & \textit{ECCV} 2024 & J & \texttimes & \texttimes & 0.7156 & 0.6304 & 0.8588 \\
        ProtoGCN~\cite{ProtoGCN} & \textit{CVPR} 2025 & J & \texttimes & \texttimes & 0.7152 & 0.6282 & 0.8808 \\
        TemPose-V~\cite{TemPose} & \textit{CVPRW} 2023 & J & \texttimes & \texttimes & 0.7154 & 0.6326 & 0.8920 \\
        \midrule
        TemPose-V~\cite{TemPose} & \textit{CVPRW} 2023 & J+B & \texttimes & \texttimes & 0.7216 & 0.6378 & 0.8952 \\
        \midrule
        TemPose-PF &  & J+B & \texttimes & \checkmark & 0.7384 & 0.6670 & 0.9090 \\
        \midrule
        TemPose-SF &  & J+B & \checkmark & \texttimes & 0.7602 & \textbf{0.6922} & 0.9196 \\
        BST-0 &  & J+B & \checkmark & \texttimes & \textbf{0.7626} & 0.6908 & \textbf{0.9206} \\
        \midrule
        TemPose-TF~\cite{TemPose} & \textit{CVPRW} 2023 & J+B & \checkmark & \checkmark & 0.7683 & 0.7027 & 0.9242 \\
        BST &  & J+B & \checkmark & \checkmark & 0.7676 & 0.6966 & 0.9236 \\
        BST-CG &  & J+B & \checkmark & \checkmark & 0.7681 & 0.7004 & 0.9252 \\
        BST-AP &  & J+B & \checkmark & \checkmark & 0.7689 & 0.7023 & 0.9260 \\
        BST-CG-AP &  & J+B & \checkmark & \checkmark & \textbf{0.7695} & \textbf{0.7043} & \textbf{0.9267} \\
        \bottomrule
    \end{tabular}
\end{table*}

\begin{table*}[htbp]
    \caption{Model comparison on ShuttleSet (35 classes) with our clipping strategy.}
    \label{tab:comparison_ShuttleSet_35_our_strategy}
    \centering
    \begin{tabular}{l|ccc@{\hspace{0.5cm}}|@{\hspace{0.5cm}}ccc}
        \toprule
        Model & Modality & SP & PP & Acc & Macro-F1 & Acc-2 \\
        \midrule
        TemPose-SF & J+B & \checkmark & \texttimes & 0.7534 & 0.6833 & 0.9194 \\
        BST-0 & J+B & \checkmark & \texttimes & \textbf{0.7655} & \textbf{0.6976} & \textbf{0.9306} \\
        \midrule
        TemPose-TF~\cite{TemPose} & J+B & \checkmark & \checkmark & 0.7580 & 0.6902 & 0.9214 \\
        BST & J+B & \checkmark & \checkmark & 0.7698 & 0.7010 & 0.9331 \\
        BST-CG & J+B & \checkmark & \checkmark & 0.7704 & 0.7026 & \textbf{0.9355} \\
        BST-AP & J+B & \checkmark & \checkmark & 0.7687 & 0.7000 & 0.9312 \\
        BST-CG-AP & J+B & \checkmark & \checkmark & \textbf{0.7710} & \textbf{0.7042} & 0.9340 \\
        \bottomrule
    \end{tabular}
\end{table*}

\begin{figure*}[htbp]
    \centering
    \includegraphics[width=\linewidth]{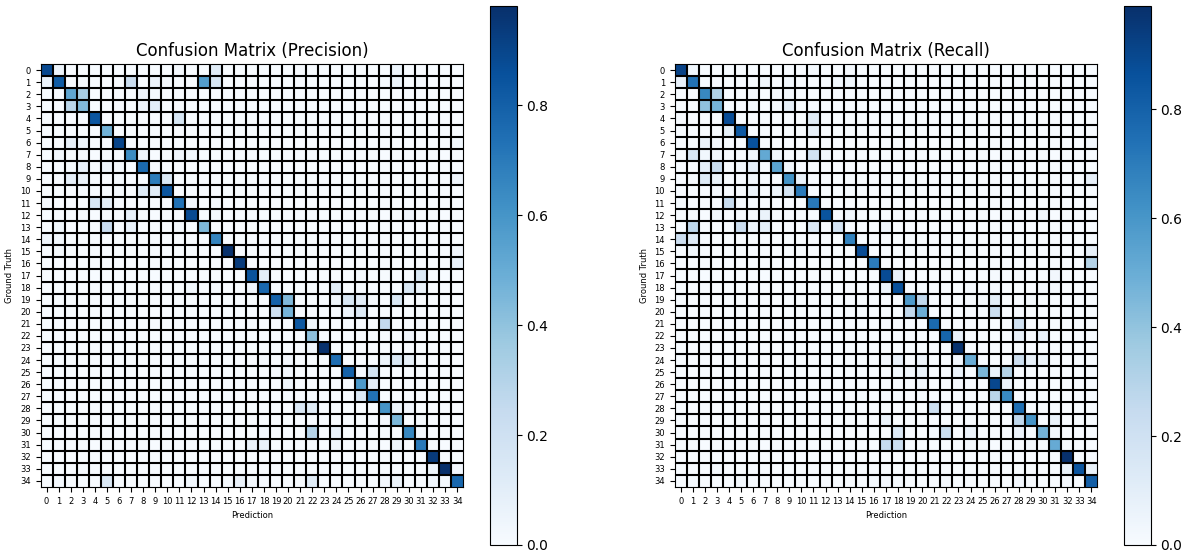}
    \caption{
        Normalized confusion matrices of BST-0 on ShuttleSet (35 classes).
        The sum of the elements in each column in the left matrix is equal to 1, and the sum of the elements in each row in the right matrix is equal to 1.
    }
    \label{fig:confusion_matrix}
\end{figure*}

\begin{table*}[htbp]
    \caption{Model comparison trained on 25\% ShuttleSet (35 classes) with fixed-width strategy.}
    \label{tab:comparison_ShuttleSet_35_reduced_training_set}
    \centering
    \begin{tabular}{l|ccc@{\hspace{0.5cm}}|@{\hspace{0.5cm}}ccc}
        \toprule
        Model & Modality & SP & PP & Acc & Macro-F1 & Acc-2 \\
        \midrule
        TemPose-SF & J+B & \checkmark & \texttimes & 0.5888 & 0.4552 & 0.7760 \\
        BST-0 & J+B & \checkmark & \texttimes & \textbf{0.6264} & \textbf{0.5136} & \textbf{0.8098} \\
        \midrule
        TemPose-TF~\cite{TemPose} & J+B & \checkmark & \checkmark & 0.5822 & 0.4372 & 0.7560 \\
        BST & J+B & \checkmark & \checkmark & 0.6323 & 0.5238 & 0.8202 \\
        BST-CG & J+B & \checkmark & \checkmark & \textbf{0.6436} & 0.5266 & \textbf{0.8234} \\
        BST-AP & J+B & \checkmark & \checkmark & 0.6374 & \textbf{0.5276} & 0.8164 \\
        BST-CG-AP & J+B & \checkmark & \checkmark & 0.6338 & 0.5204 & 0.8218 \\
        \bottomrule
    \end{tabular}
\end{table*}

\hypertarget{discussion_cls}{}
\subsection{Classifying Difficulty on ShuttleSet}
\label{subsec:classifying_difficulty}

At the beginning of this work, we ran the experiments on ShuttleSet~\cite{ShuttleSet} (35 classes), but the accuracy of it was not high enough, as shown in \cref{tab:comparison_ShuttleSet_35_fixed_width_strategy} and \cref{tab:comparison_ShuttleSet_35_our_strategy}.
We suspect that this is primarily due to the excessive granularity of certain categories.
The normalized confusion matrices of BST-0 are shown in \cref{fig:confusion_matrix}.
The 0th to 16th strokes are executed by the top player, the 17th to 33rd strokes are hit by the bottom player, and the 34th stroke belongs to the "none" category.
Observing on both confusion matrices, the model struggles to differentiate between similar stroke-types, such as the 2nd and 3rd strokes (top smash and top wrist smash) and the 19th and 20th strokes (bottom smash and bottom wrist smash).
A closer inspection reveals that the model often misclassifies the 1st stroke (top return net) as the 13th stroke (top defensive return drive) that is shown in the left matrix.
Additionally viewing the diagonal elements in the right matrix, the 13th stroke (top defensive return drive) has the worst recall.
These results suggest that the model has difficulty distinguishing between subtle variations in the strokes with similar hitting characteristics.
We observed the same phenomenon when employing other models as well.

For a deeper look into the results between \cref{tab:comparison_ShuttleSet_35_fixed_width_strategy} and \cref{tab:comparison_ShuttleSet_35_our_strategy}, we can observe two phenomena.
First, the performance of TemPose drops when using our clipping strategy.
Our clipping strategy creates variable sequence lengths due to the fluctuating match pace.
TemPose struggles with these, possibly due to limitations in its Interactional Transformer, which is designed to process \textit{people} as a sequence.
(We have mentioned this in the last paragraph of \cref{subsubsec:BST-0}.)
The other phenomenon is that when comparing the results of BSTs themselves, the differences between BSTs (fixed-width vs. our clipping strategy) from BST to BST-CG-AP are [\mbox{+0.22\%}, \mbox{+0.44\%}, \mbox{+0.95\%}], [\mbox{+0.23\%}, \mbox{+0.22\%}, \mbox{+1.03\%}], [\mbox{-0.02\%}, \mbox{-0.23\%}, \mbox{+0.52\%}], [\mbox{+0.15\%}, \mbox{-0.01\%}, \mbox{+0.73\%}].
Notably, 2 of the 3 observed performance drops occur with BST-AP.
This is probably because BST-AP's pathways, prior to the MLP head, are limited to two branches (information from the blue player and from the green player), lacking the overall trajectory information.
Furthermore, our clipping strategy primarily focuses on shuttlecock trajectory information.
Consequently, the performance drop is more pronounced for BST-AP than for BST-CG-AP.
Despite a modest drop on 3 out of 12 metrics, BSTs still benefit from our clipping strategy, especially with high class granularity, where \mbox{Acc-2} becomes more important.
Moreover, compared to the best of TemPose in \cref{tab:comparison_ShuttleSet_35_fixed_width_strategy}, BSTs (with our strategy) outperform it on 9 out of 12 metrics, with one loss being only \mbox{-0.01\%}.
Lastly, our backbone BST-0, even without PP input but utilizing our clipping strategy, still beats the best of TemPose on \mbox{Acc-2} (0.9306 vs. 0.9242).

\cref{tab:comparison_ShuttleSet_35_reduced_training_set} shows the results of models trained on only (each class) 25\% of the training set and tested on the same testing set.
This indicates that our BST architecture still demonstrates superior generalization capability compared to TemPose when the training dataset is small and reference information is limited, even with excessive class granularity.

\hypertarget{discussion_2D_vs_3D}{}
\subsection{Player 2D Joints vs. 3D Joints}

\begin{figure*}[htbp]
    \centering
    \begin{subfigure}{0.24\linewidth}
        \centering
        \includegraphics[width=0.95\linewidth]{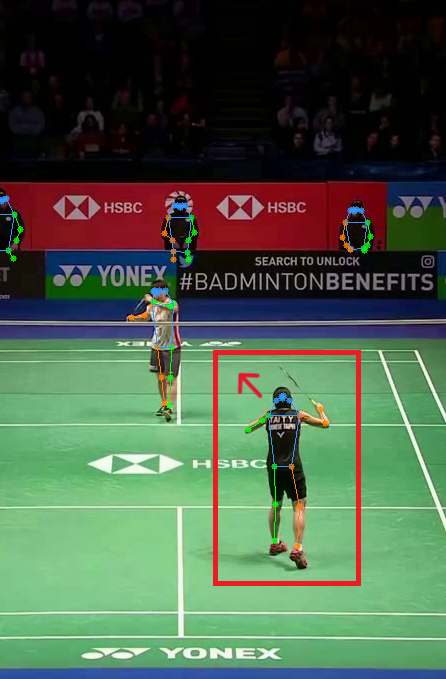}
        \caption{2D pose}
        \label{fig:2D_vs_3D-a}
    \end{subfigure}
    \hfill
    \begin{subfigure}{0.24\linewidth}
        \centering
        \includegraphics[width=0.95\linewidth]{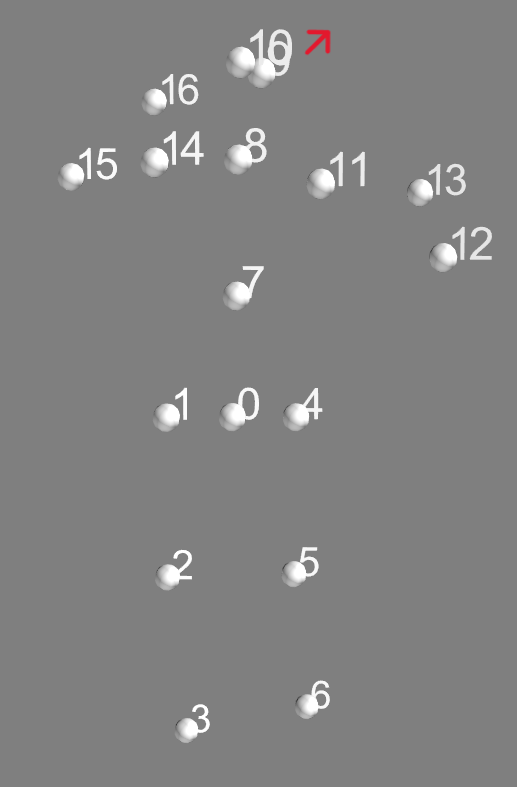}
        \caption{Rear view of the 3D pose}
        \label{fig:2D_vs_3D-b}
    \end{subfigure}
    \hfill
    \begin{subfigure}{0.24\linewidth}
        \centering
        \includegraphics[width=0.95\linewidth]{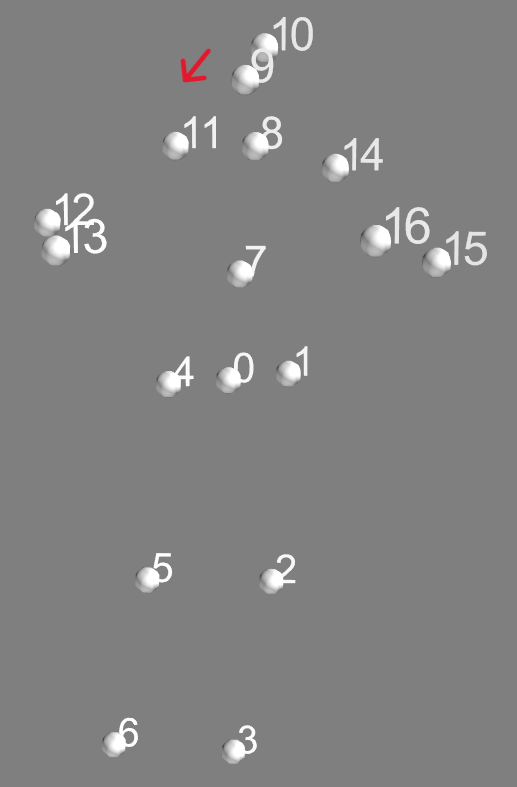}
        \caption{Front view of the 3D pose}
        \label{fig:2D_vs_3D-c}
    \end{subfigure}
    \hfill
    \begin{subfigure}{0.24\linewidth}
        \centering
        \includegraphics[width=0.95\linewidth]{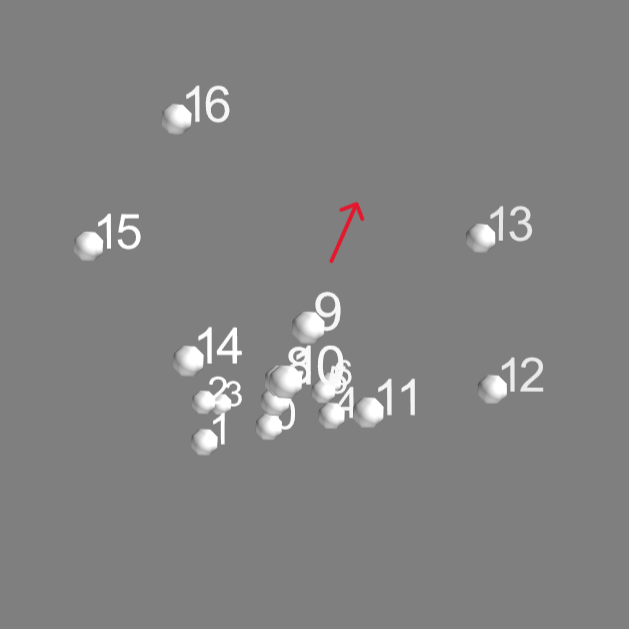}
        \caption{Top view of the 3D pose}
        \label{fig:2D_vs_3D-d}
    \end{subfigure}
    
    \caption{
        Example of a 2D pose vs. 3D pose in a badminton video frame.
        The ninth joint represents the human nose.
        The red arrows indicate the directions the bottom player was facing.
        We can see the bottom player was facing her opponent in fact, but the 3D pose exhibits a clear error in the facing direction.
        We suspect that this 3D model will often assume that if both forearms are pointing roughly forward, it will lead the model to assume that the face is also pointing in that direction.
        (The visualizations in (b), (c), and (d) are generated by the Mayavi~\cite{Mayavi} python package.)
    }
    \label{fig:2D_vs_3D}
\end{figure*}

\begin{table*}[htbp]
    \caption{
        Model with 2D vs. 3D poses comparison on ShuttleSet (35 classes) with fixed-width clipping strategy.
        The value on the left side of the slash is the result of the model with 2D poses, and the right side is the result of the model with 3D poses.
    }
    \label{tab:2D_vs_3D}
    \centering
    \begin{tabular}{l|@{\hspace{0.5cm}}ccc}
        \toprule
        Model & Accuracy & Macro-F1 & Accuracy-2 \\
        \midrule
        TemPose-SF & \textbf{0.7602} / 0.7432 & \textbf{0.6922} / 0.6670 & \textbf{0.9196} / 0.9062 \\
        BST-0 & \textbf{0.7626} / 0.7456 & \textbf{0.6908} / 0.6696 & \textbf{0.9206} / 0.9102 \\
        \midrule
        TemPose-TF~\cite{TemPose} & \textbf{0.7683} / 0.7579 & \textbf{0.7027} / 0.6866 & \textbf{0.9242} / 0.9176 \\
        BST-CG-AP & \textbf{0.7695} / 0.7525 & \textbf{0.7043} / 0.6760 & \textbf{0.9267} / 0.9181 \\
        \bottomrule
    \end{tabular}
\end{table*}

While we explored using 3D player joints as input to our model, the performance was inferior to that achieved with 2D joints.
We attribute this to the fact that the HPE model is trained on a general HPE dataset, encompassing a broader range of poses than those specific to badminton players.
Additionally, the 3D joints are estimated from the 2D joints, which may further contribute to a bias towards general human poses, rather than the specialized poses of badminton players.
We can see an example of this in \cref{fig:2D_vs_3D}.
This issue would mislead the model to predict the wrong action, since the 3D pose is not accurate enough to represent the player's pose in the badminton game, as shown in \cref{tab:2D_vs_3D}.

\hypertarget{discussion_future_work}{}
\subsection{Limitations and Future Work}

Our method has an obvious limitation: it relies on accurate hit frame detection and shuttlecock trajectory tracking models.
Comparing between \cref{tab:comparison_ShuttleSet_25_fixed_width_strategy} and \cref{tab:comparison_ShuttleSet_25_our_strategy} shows that too much unrelated information or too little correct information to the target stroke in the inputs affects the performance, which implies the hit frame accuracy also affects the performance.
Fortunately, these issues are expected to be resolved in the future, similar to how the transition from HAR to SAR was facilitated by advanced HPE models, which enabled SAR to perform more effectively.
Moreover, the existing 2D shuttlecock trajectory tracking models are quite mature, and we believe that there will be more 3D tracking models in the future.

Although this paper primarily focuses on stroke classification in singles badminton, and also experiments on tennis, the potential applicability of our method to other racket sports, such as table tennis, or even doubles, is worth exploring.

\hypertarget{clipping_details}{}
\section{Details in Clipping}

\subsection{Source of the Hit Frame and Rally Information}
Since the clipping strategies are predefined in BadmintonDB~\cite{BadmintonDB} and TenniSet~\cite{TenniSet}, and the hit frame and rally information are provided in ShuttleSet~\cite{ShuttleSet}, we don't need to detect hit frames and rallies by ourselves.
If these were not provided, we could use the HitNet model proposed in MonoTrack~\cite{MonoTrack} to detect hit frames.
To detect rallies, we note that time intervals between successive rallies are in general much larger than the time differences between two typical successive hit frames in one rally.
For example, the time difference between the last hit frame number of a rally and the first hit frame number of the next rally must be larger than that between the two consecutive hit frames in the same rally, which can be presented as $h^{(i+1)}_1 - h^{(i)}_{n^{(i)}} > h^{(k)}_{j+1} - h^{(k)}_j, \forall\ 1\le k \le r$, where $r$ is the total number of rallies in the match video.

\subsection{Practical Settings}
In practice, we set some limits on $\hat{h}^{(i)}_{j-1}$ and $\hat{h}^{(i)}_{j+1}$ in our strategy to prevent the model from accessing irrelevant information caused by remaining unclean data.
Specifically, the actual time distance of these two frame numbers from the target frame number does not exceed 1.5 seconds (before adding $\epsilon$).
Additionally, $t$ is set to the half of the FPS of the video representing 0.5 seconds interval, and $\epsilon$ is set to $\frac{t}{2}$.

\hypertarget{training_settings}{}
\section{Training Settings}

The implementation is based on the PyTorch framework.
All downstream classification models in this paper were trained from scratch using cross-entropy loss.
Moreover, the compared models are derived from their official public implementations.
During training, we applied random shift data augmentation with a probability of $0.3$, where the shift value was uniformly sampled from $[-0.3, 0.3)$.
We chose AdamW as the optimizer and cosine annealing with warm-up learning rate scheduler to adjust the learning rate.
We validated the model on the validation set after each epoch and saved the model with the best macro-F1 score on the validation set.
The hyperparameters used in our experiments are listed below:

\begin{itemize}[leftmargin=2em]
    \item \textbf{n\_epochs}: $1600$
    \item \textbf{early\_stop\_n\_epochs}: $300$
    \item \textbf{batch\_size}: $128$
    \item \textbf{learning\_rate}:
    \begin{itemize}[$\scriptstyle\circ$]
        \item $5\times 10^{-2}$ on BlockGCN
        \item $1\times 10^{-3}$ on ST-GCN, ProtoGCN, SkateFormer
        \item $5\times 10^{-4}$ on other models
    \end{itemize}
    \item \textbf{cosine\_annealing\_num\_cycles}: $0.25$
    \item \textbf{warm\_up\_step}: $400$
    \item \textbf{weight\_decay}:
    \begin{itemize}[$\scriptstyle\circ$]
        \item $5\times 10^{-4}$ on SkateFormer partial weights
        \item $1\times 10^{-2}$ on other models
    \end{itemize}
    \item \textbf{label\_smoothing}: $0.1$
    \item \textbf{sequence\_length}:
    \begin{itemize}[$\scriptstyle\circ$]
        \item $30$ on ShuttleSet with fixed-width strategy
        \item $100$ on ShuttleSet with our strategy
        \item $72$ on BadmintonDB
        \item $100$ on TenniSet
    \end{itemize}
    \item \textbf{n\_classes}:
    \begin{itemize}[$\scriptstyle\circ$]
        \item $35$ on ShuttleSet (excessive granularity)
        \item $25$ on ShuttleSet (merged)
        \item $18$ on BadmintonDB
        \item $6$ on TenniSet
    \end{itemize}
\end{itemize}

\hypertarget{datasets_details}{}
\section{Details of the Datasets}

\textbf{ShuttleSet (excessive granularity).}\quad
The original dataset consists of 19 distinct stroke categories, including a "none" type.
One category, driven flight, containing only 52 strokes across the entire dataset was merged into the "none" category by us. \\
Merged types: \vspace{0.3em}\\
\begin{tabular}{@{\hspace{1em}}cr@{\ $\Rightarrow$\ }l}
    $\scriptstyle\bullet$ & "driven flight" & "none" \\
\end{tabular} \vspace{0.3em}\\
Additionally, since the dataset does not differentiate between strokes played by the top or bottom player within each category, the number of stroke categories (except for "none") was doubled.
This adjustment resulted in a total of 35 categories, including "none", as shown in \cref{tab:classes_shuttleset}.

The court information is provided in the dataset, so we don't need to detect it by ourselves.
If there were less than two people in the court, we cleared the information (poses and shuttlecock trajectory) of that frame to zero, since that frame was definitely not in a standard camera perspective.

\begin{table*}[htbp]
    \caption{
        Classes in ShuttleSet (excessive granularity).
    }
    \label{tab:classes_shuttleset}
    \centering
    \begin{tabular}{r@{\hspace{0.5cm}}|@{\hspace{0.5cm}}rrrrcll}
        \toprule
        Class ID & \#Train & \#Val & \#Test & \#Total & Player & 中文類別名稱 & Class Name \\
        \midrule
        0 & 2402 & 357 & 223 & 2982 & Top & 放小球 & net shot \\
        1 & 1257 & 179 & 182 & 1618 & Top & 擋小球 & return net \\
        2 & 980 & 185 & 122 & 1287 & Top & 殺球 & smash \\
        3 & 605 & 120 & 106 & 831 & Top & 點扣 & wrist smash \\
        4 & 1858 & 266 & 219 & 2343 & Top & 挑球 & lob \\
        5 & 96 & 15 & 12 & 123 & Top & 防守回挑 & defensive return lob \\
        6 & 925 & 197 & 187 & 1309 & Top & 長球 & clear \\
        7 & 258 & 28 & 33 & 319 & Top & 平球 & drive \\
        8 & 172 & 27 & 35 & 234 & Top & 後場抽平球 & back-court drive \\
        9 & 782 & 153 & 101 & 1036 & Top & 切球 & drop \\
        10 & 463 & 94 & 58 & 615 & Top & 過渡切球 & passive drop \\
        11 & 1005 & 156 & 156 & 1317 & Top & 推球 & push \\
        12 & 203 & 33 & 34 & 270 & Top & 撲球 & rush \\
        13 & 96 & 26 & 22 & 144 & Top & 防守回抽 & defensive return drive \\
        14 & 503 & 60 & 43 & 606 & Top & 勾球 & cross-court net shot \\
        15 & 733 & 84 & 87 & 904 & Top & 發短球 & short service \\
        16 & 97 & 19 & 43 & 159 & Top & 發長球 & long service \\
        \midrule
        17 & 2282 & 338 & 222 & 2842 & Bottom & 放小球 & net shot \\
        18 & 1381 & 207 & 168 & 1756 & Bottom & 擋小球 & return net \\
        19 & 780 & 152 & 143 & 1075 & Bottom & 殺球 & smash \\
        20 & 589 & 58 & 81 & 728 & Bottom & 點扣 & wrist smash \\
        21 & 1974 & 347 & 215 & 2536 & Bottom & 挑球 & lob \\
        22 & 109 & 32 & 14 & 155 & Bottom & 防守回挑 & defensive return lob \\
        23 & 946 & 214 & 192 & 1352 & Bottom & 長球 & clear \\
        24 & 263 & 30 & 42 & 335 & Bottom & 平球 & drive \\
        25 & 132 & 36 & 33 & 201 & Bottom & 後場抽平球 & back-court drive \\
        26 & 741 & 112 & 90 & 943 & Bottom & 切球 & drop \\
        27 & 452 & 65 & 66 & 583 & Bottom & 過渡切球 & passive drop \\
        28 & 1030 & 169 & 136 & 1335 & Bottom & 推球 & push \\
        29 & 155 & 31 & 15 & 201 & Bottom & 撲球 & rush \\
        30 & 170 & 32 & 36 & 238 & Bottom & 防守回抽 & defensive return drive \\
        31 & 498 & 61 & 61 & 620 & Bottom & 勾球 & cross-court net shot \\
        32 & 763 & 90 & 101 & 954 & Bottom & 發短球 & short service \\
        33 & 114 & 27 & 59 & 200 & Bottom & 發長球 & long service \\
        \midrule
        34 & 927 & 241 & 162 & 1330 & - & 未知球種 & none \\
        \midrule
        \#Sum & 25741 & 4241 & 3499 & 33481 & - & - & - \\
        \bottomrule
    \end{tabular}
\end{table*}

\vspace{0.8em}

\noindent\textbf{ShuttleSet (merged).}\quad
Everything is the same as the excessive granularity version, except for the following merged categories: \vspace{0.3em}\\
\begin{tabular}{@{\hspace{1em}}cr@{\ $\Rightarrow$\ }l}
    $\scriptstyle\bullet$ & "wrist smash" & "smash" \\
    $\scriptstyle\bullet$ & "defensive return lob" & "lob" \\
    $\scriptstyle\bullet$ & "driven flight" & "drive" \\
    $\scriptstyle\bullet$ & "back-court drive" & "drive" \\
    $\scriptstyle\bullet$ & "defensive return drive" & "drive" \\
    $\scriptstyle\bullet$ & "passive drop" & "drop" \\
\end{tabular} \vspace{0.3em}\\
Thus, there are 25 classes in this dataset as shown in \cref{tab:classes_shuttleset_merged}.

\begin{table*}[htbp]
    \caption{
        Classes in ShuttleSet (merged).
    }
    \label{tab:classes_shuttleset_merged}
    \centering
    \begin{tabular}{r@{\hspace{0.5cm}}|@{\hspace{0.5cm}}rrrrcll}
        \toprule
        Class ID & \#Train & \#Val & \#Test & \#Total & Player & 中文類別名稱 & Class Name \\
        \midrule
        0 & 875 & 241 & 162 & 1278 & - & 未知球種 & none \\
        \midrule
        1 & 2402 & 357 & 223 & 2982 & Top & 放小球 & net shot \\
        2 & 1257 & 179 & 182 & 1618 & Top & 擋小球 & return net \\
        3 & 1585 & 305 & 228 & 2118 & Top & 殺球 & smash \\
        4 & 1954 & 281 & 231 & 2466 & Top & 挑球 & lob \\
        5 & 925 & 197 & 187 & 1309 & Top & 長球 & clear \\
        6 & 548 & 81 & 90 & 719 & Top & 平球 & drive \\
        7 & 1245 & 247 & 159 & 1651 & Top & 切球 & drop \\
        8 & 1005 & 156 & 156 & 1317 & Top & 推球 & push \\
        9 & 203 & 33 & 34 & 270 & Top & 撲球 & rush \\
        10 & 503 & 60 & 43 & 606 & Top & 勾球 & cross-court net shot \\
        11 & 733 & 84 & 87 & 904 & Top & 發短球 & short service \\
        12 & 97 & 19 & 43 & 159 & Top & 發長球 & long service \\
        \midrule
        13 & 2282 & 338 & 222 & 2842 & Bottom & 放小球 & net shot \\
        14 & 1381 & 207 & 168 & 1756 & Bottom & 擋小球 & return net \\
        15 & 1369 & 210 & 224 & 1803 & Bottom & 殺球 & smash \\
        16 & 2083 & 379 & 229 & 2691 & Bottom & 挑球 & lob \\
        17 & 946 & 214 & 192 & 1352 & Bottom & 長球 & clear \\
        18 & 595 & 98 & 111 & 804 & Bottom & 平球 & drive \\
        19 & 1193 & 177 & 156 & 1526 & Bottom & 切球 & drop \\
        20 & 1030 & 169 & 136 & 1335 & Bottom & 推球 & push \\
        21 & 155 & 31 & 15 & 201 & Bottom & 撲球 & rush \\
        22 & 498 & 61 & 61 & 620 & Bottom & 勾球 & cross-court net shot \\
        23 & 763 & 90 & 101 & 954 & Bottom & 發短球 & short service \\
        24 & 114 & 27 & 59 & 200 & Bottom & 發長球 & long service \\
        \midrule
        \#Sum & 25741 & 4241 & 3499 & 33481 & - & - & - \\
        \bottomrule
    \end{tabular}
\end{table*}

\vspace{0.8em}

\noindent\textbf{BadmintonDB.}\quad
Merged types: \vspace{0.3em}\\
\begin{tabular}{@{\hspace{1em}}cr@{\ $\Rightarrow$\ }l}
    $\scriptstyle\bullet$ & "Flick-Serve" & "Serve" \\
\end{tabular} \vspace{0.3em}\\
Thus, there are 18 classes in this dataset as shown in \cref{tab:classes_badDB}.

In court detection (using MonoTrack~\cite{MonoTrack}), since there is only one standard camera perspective in a badminton match, it only needs to be performed once per match video, on a single frame that meets this criterion.
If the court detection fails on that frame, actually, it will likely fail on all frames of that match video.
We marked the four corners of the court manually in this case.
If there were less than two people in the court, we conducted the same operation as in ShuttleSet.

We used duplicated data to balance the classes during training.

The reason for SkateFormer~\cite{SkateFormer} not in the comparison table is that setting a proper sequence length for it is not an easy task.

\begin{table*}[htbp]
    \caption{
        Classes in BadmintonDB.
    }
    \label{tab:classes_badDB}
    \centering
    \begin{tabular}{r@{\hspace{0.5cm}}|@{\hspace{0.5cm}}rrrrcl}
        \toprule
        Class ID & \#Train & \#Val & \#Test & \#Total & Player & Class Name \\
        \midrule
        0 & 497 & 62 & 63 & 622 & Bottom & Block \\
        1 & 263 & 32 & 34 & 329 & Bottom & Clear \\
        2 & 60 & 7 & 9 & 76 & Bottom & Drive \\
        3 & 260 & 32 & 33 & 325 & Bottom & Dropshot \\
        4 & 28 & 3 & 5 & 36 & Bottom & Net-Kill \\
        5 & 734 & 91 & 93 & 918 & Bottom & Net-Lift \\
        6 & 663 & 82 & 84 & 829 & Bottom & Net-Shot \\
        7 & 179 & 22 & 23 & 224 & Bottom & Serve \\
        8 & 392 & 49 & 49 & 490 & Bottom & Smash \\
        \midrule
        9 & 458 & 57 & 58 & 573 & Top & Block \\
        10 & 203 & 25 & 26 & 254 & Top & Clear \\
        11 & 74 & 9 & 10 & 93 & Top & Drive \\
        12 & 289 & 36 & 37 & 362 & Top & Dropshot \\
        13 & 52 & 6 & 8 & 66 & Top & Net-Kill \\
        14 & 712 & 89 & 89 & 890 & Top & Net-Lift \\
        15 & 639 & 79 & 81 & 799 & Top & Net-Shot \\
        16 & 157 & 19 & 21 & 197 & Top & Serve \\
        17 & 460 & 57 & 58 & 575 & Top & Smash \\
        \midrule
        \#Sum & 6120 & 757 & 781 & 7658 & - & - \\
        \bottomrule
    \end{tabular}
\end{table*}

\vspace{0.8em}

\noindent\textbf{TenniSet.}\quad
This dataset contains 6 classes as shown in \cref{tab:classes_tenniSet}.

In court detection (using TennisCourtDetector~\cite{TennisCourtDetector}), it was performed on each frame, as the camera perspective often shifts slightly left or right during a rally.

When determining who the two players are among the people in a frame, since the two tennis players are often outside the court lines, instead of the situations in badminton, we choose the person(s) closest to the top or bottom side of the court line(s), if there were less than two people inside the court lines.
If there were still less than two people in the set we picked, we cleared the information (poses and ball trajectory) of that frame to zero, since the court detection was probably failed in this case (not in a standard camera perspective or model's failure).

In ball trajectory tracking, we used TrackNetV3~\cite{TrackNetV3-Attention} (with attention), which was originally trained on badminton videos, for convenience.
Although the results were not perfect, often missing the ball during bounces, the downstream model was still able to learn useful information.

The reason for other three models not in the comparison table is that they take too long time and too many resources to train on a long sequence length (\eg 100).

\begin{table*}[htbp]
    \caption{
        Classes in TenniSet.
        "Far" means "Top" and "Near" means "Bottom".
        "Left" means that the ball was on the left side of the player who hit it in the camera perspective (not the player's perspective).
        "Right" means the opposite.
    }
    \label{tab:classes_tenniSet}
    \centering
    \begin{tabular}{r@{\hspace{0.5cm}}|@{\hspace{0.5cm}}rrrrl}
        \toprule
        Class ID & \#Train & \#Val & \#Test & \#Total & Class Name \\
        \midrule
        0 & 432 & 33 & 135 & 600 & HFL (Hit Far Left) \\
        1 & 474 & 37 & 145 & 656 & HFR (Hit Far Right) \\
        2 & 514 & 37 & 119 & 670 & HNL (Hit Near Left) \\
        3 & 448 & 33 & 144 & 625 & HNR (Hit Near Right) \\
        \midrule
        4 & 412 & 30 & 90 & 532 & SF (Serve Far) \\
        5 & 339 & 27 & 120 & 486 & SN (Serve Near) \\
        \midrule
        \#Sum & 2619 & 197 & 753 & 3569 & - \\
        \bottomrule
    \end{tabular}
\end{table*}

\end{document}